\documentclass[runningheads]{llncs}
\usepackage{graphicx}

\usepackage{tikz}
\usepackage{comment}
\usepackage{amsmath,amssymb} 
\usepackage{color}

\usepackage{phonetic}
\usepackage{cite}
\usepackage{enumitem}
\usepackage[percent]{overpic}
\usepackage{multirow}
\usepackage{xcolor}
\usepackage[linesnumbered]{algorithm2e}

\usepackage[pagebackref=true,breaklinks=true,letterpaper=true,colorlinks=true,linkcolor=black,citecolor=black,bookmarks=false]{hyperref}

\begin{document}
\pagestyle{headings}
\mainmatter
\def\ECCVSubNumber{4241} 

\def\httilde{\mbox{\tt\raisebox{-.5ex}{\symbol{126}}}}

\newcommand{\Sim}{\mathord{\sim}}

\newcommand{\Paragraph}[1]{\vspace{1.25mm} \noindent \textbf{#1} \hspace{0mm}}
\newcommand{\Section}[1]{\vspace{-1.85mm} \section{#1} \vspace{-1.85mm}}
\newcommand{\SubSection}[1]{\vspace{-1.75mm} \subsection{#1} \vspace{-1.75mm}}
\newcommand{\SubSubSection}[1]{\vspace{-1.75mm} \subsubsection{#1} \vspace{-1.75mm}}

\newcommand{\xdownarrow}[1]{%
  {\left\downarrow\vbox to #1{}\right.\kern-\nulldelimiterspace}
}

\newcommand\two{\text{2D}}
\newcommand\three{\text{3D}}

\newcommand{\bernt}[1]{\textcolor[rgb]{0.08, 0.38, 0.74}{\textbf{Bernt: #1}}}
\newcommand{\GPM}[1]{\textcolor[rgb]{0.08, 0.8, 0.0}{\textbf{Gerard: #1}}}

\title{Kinematic 3D Object Detection in Monocular Video}

%
\author{Garrick Brazil\inst{1} \and
Gerard Pons-Moll\inst{2} \and
Xiaoming Liu\inst{1} \and Bernt Schiele\inst{2}}
\authorrunning{G. Brazil et al.}
%
\institute{Michigan State University, Computer Science $\&$ Engineering \\ \and
Max Planck Institute for Informatics, Saarland Informatics Campus\\
\inst{1}~\email{\{brazilga,~liuxm\}@msu.edu},~~\inst{2}~\email{\{gpons,~schiele\}@mpi-inf.mpg.de}
}

\maketitle

\begin{abstract}
Perceiving the physical world in 3D is fundamental for self-driving applications. 
Although temporal motion is an invaluable resource to human vision for detection, tracking, and depth perception, such features have not been thoroughly utilized in modern 3D object detectors.\linebreak
In this work, we propose a novel method for monocular video-based 3D object detection which leverages kinematic motion to extract scene dynamics and improve localization accuracy. 
We first propose a novel decomposition of object orientation and a self-balancing 3D confidence.\linebreak
We show that both components are critical to enable our kinematic model to work effectively.
Collectively, using only a single model, we efficiently leverage 3D kinematics from monocular videos to improve the overall localization precision in 3D object detection while also producing useful by-products of scene dynamics (ego-motion and per-object velocity). \linebreak
We achieve state-of-the-art performance on monocular 3D object detection and the Bird's Eye View tasks within the KITTI self-driving dataset. 
\end{abstract}

\Section{Introduction}

\begin{figure*}[t]
\begin{center}
   \includegraphics[width=\linewidth]{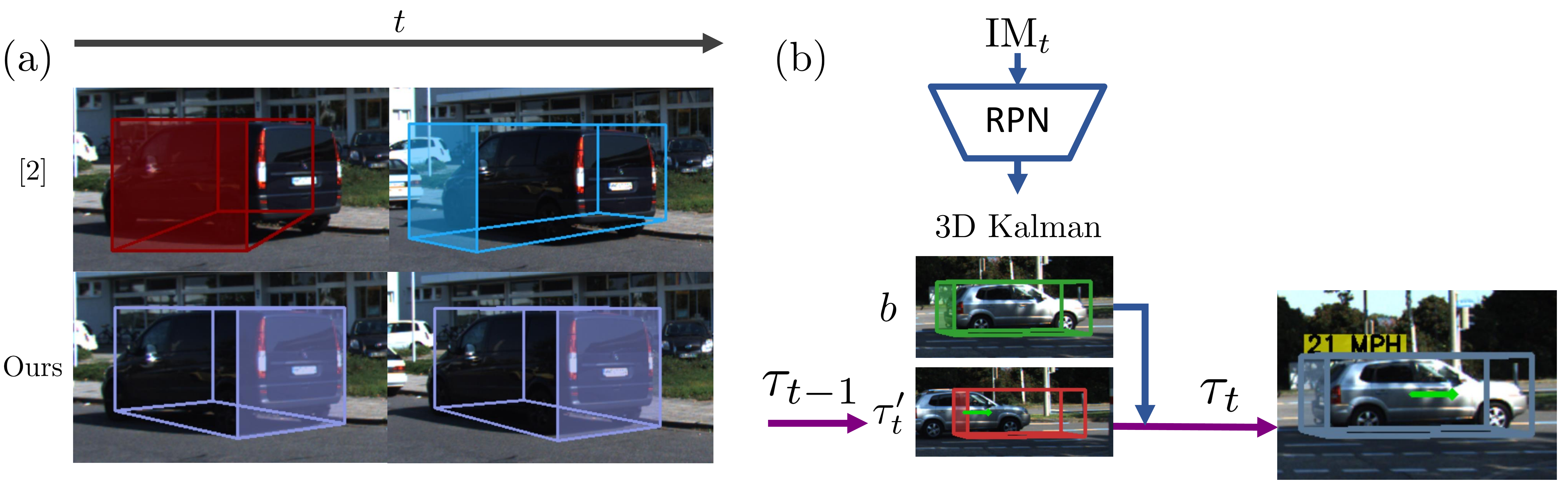}
\vspace{-5mm}
      \caption{
Single-frame 3D detection~\cite{brazil2019m3d} often has unstable estimation through time (a), while our video-based method (b) is more robust by leveraging \textbf{kinematic motion} via a 3D Kalman Filter to fuse forecasted tracks $\tau'_t$ and measurements $b$ into $\tau_t$.
}

\label{fig:teaser}
\end{center}\vspace{-7mm}
\end{figure*}
The detection of foreground objects is among the most critical requirements to facilitate self-driving applications~\cite{dollar2012pedestrian,illuminating-pedestrians-via-simultaneous-detection-segmentation,pedestrian-detection-with-autoregressive-network-phases}. 
Recently, 3D object detection has made significant progress~\cite{chen2019fast, lang2019pointpillars, liang2018deep, liang2019multi, shi2019pointrcnn, yang2019std}, even while using only a monocular camera~\cite{brazil2019m3d, ku2019monocular, li2019gs3d, liu2019deep, ma2019accurate, manhardt2019roi, simonelli2019disentangling, wang2019pseudo}. 
Such works primarily look at the problem from the perspective of \textit{single} frames, ignoring useful temporal cues and constraints.

Computer vision cherishes inverse problems, \emph{e.g.}, recovering the 3D physical motion of objects from monocular videos.
Motion information such as object velocity in the metric space is highly desirable for the path planning of self-driving. 
However, single image-based 3D object detection can not directly estimate physical motion, without relying on additional tracking modules.
Therefore, \textit{video-based 3D object detection} would be a sensible choice to recover such motion information.
Furthermore, without modeling the physical motion, image-based 3D object detectors are naturally more likely to suffer from erratic and unnatural changes through time in orientation and localization (as exemplified in Fig.~\ref{fig:teaser}(a)). 
Therefore, we aim to build a novel video-based 3D object detector which is able to provide \textit{accurate} and \textit{smooth} 3D object detection with per-object \textit{velocity},  while also prioritizing a \textit{compact} and \textit{efficient} model overall.

Yet, designing an effective video-based 3D object detector has challenges.
Firstly, motion which occurs in real-world scenes can come from a variety of sources such as the camera atop of an autonomous vehicle or robot, and/or from the scene objects themselves --- for which most of the safety-critical objects (car, pedestrian, cyclist~\cite{Geiger2012CVPR}) are typically \textit{dynamic}.
Moreover, using video inherently involves an increase in data consumption which introduces practical challenges for training and/or inference including with memory or redundant processing. 

To address such challenges, we propose a novel framework to integrate a 3D Kalman filter into a 3D detection system.
We find Kalman is an ideal candidate for three critical reasons: (1) it allows for use of real-world motion models to serve as a strong prior on object dynamics, (2) it is inherently efficient due to its recursive nature and general absence of parameters, (3) the resultant behavior is explainable and provides useful by-products such as the object velocity.

Furthermore, we observe that objects predominantly move in the direction indicated by their orientation. 
Fortunately, the benefit of Kalman allows us to integrate this real-world constraint into the motion model as a compact scalar velocity. 
Such a constraint helps maintain the consistency of velocity over time and enables the Kalman motion forecasting and fusion to perform accurately. 

However, a model restricted to only move in the direction of its orientation has an obvious flaw --- what if the orientation itself is \textit{inaccurate}?
We therefore propose a novel reformulation of orientation in favor of accuracy and stability. \linebreak
We find that our orientation improves the 3D localization accuracy by a margin of $2.39\%$ and reduces the orientation error by $\approx 20\%$, which collectively help enable the proposed Kalman to function more effectively.

A notorious challenge of using Kalman comes in the form of \textit{uncertainty}, which is conventionally~\cite{welch1995introduction} assumed to be known and static, \textit{e.g.}, from a sensor.
However, 3D objects in video are intuitively dependent on more complex factors of image features and cannot necessarily be treated like a sensor measurement. 
For a better understanding of 3D uncertainty, we propose a 3D self-balancing confidence loss. 
We show that our proposed confidence has higher correlation with the 3D localization performance compared to the typical classification probability, which is commonly used in detection~\cite{ren2015faster, redmon2016you}. 

To complete the full understanding of the scene motion, we elect to estimate the ego-motion of the capturing camera itself.
Hence, we further narrow the work of Kalman to account for only the \textit{object's} motion. 
Collectively, our proposed framework is able to model important scene dynamics, both ego-motion and per-object velocity, and more precisely detect 3D objects in videos using a stabilized orientation and 3D confidence estimation. 
We demonstrate that our method achieves state-of-the-art (SOTA) performance on monocular 3D Object Detection and Bird's Eye View (BEV) tasks in the KITTI dataset~\cite{Geiger2012CVPR}.

\noindent In summary, our contributions are as follows:
\begin{itemize}[noitemsep,topsep=1mm,label=$\bullet$]
\setlength\itemsep{1mm}
\item We propose a monocular video-based 3D object detector, leveraging realistic motion constraints with an integrated ego-motion and a 3D Kalman filter.
\item We propose to reformulate orientation into axis, heading and offset along with a self-balancing 3D localization loss to facilitate the stability necessary for the proposed Kalman filter to perform more effectively.
\item Overall, using only a \textit{single model} our framework develops a comprehensive 3D scene understanding including object cuboids, orientation, velocity, object motion, uncertainty, and ego-motion, as detailed in Fig.~\ref{fig:teaser} and \ref{fig:overview}. 
\item We achieve a new SOTA performance on monocular 3D object detection and BEV tasks using comprehensive metrics within the KITTI dataset. 
\end{itemize}


\Section{Related Work}
We first provide context of our novelties from the perspective of monocular 3D object detection (Sec.~\ref{sec:related_3d}) with attention to orientation and uncertainty estimation. We next discuss and contrast with video-based object detection (Sec.~\ref{sec:related_video}).

\vspace{-1mm}
\SubSection{Monocular 3D Object Detection}
\label{sec:related_3d}
Monocular 3D object detection has made significant progress~\cite{brazil2019m3d, chen20153d, chen2016monocular, ku2019monocular, liu2019deep, li2019gs3d, ma2019accurate, manhardt2019roi, mousavian20173d, simonelli2019disentangling, wang2020task, xu2018multi}.
Early methods such as \cite{chen20153d} began by generating 3D object proposals along a ground plane using object priors and estimated point clouds, culminating in an energy minimization approach. 
\cite{chen2016monocular, ku2019monocular, xu2018multi} utilize additional domains of semantic segmentation, object priors, and estimated depth to improve the localization. 
Similarly, \cite{ma2019accurate, wang2019pseudo} create a pseudo-LiDAR map using SOTA depth estimator~\cite{chang2018pyramid, fu2018deep, felzenszwalb2010object}, which is respectfully passed into detection subnetworks or LiDAR-based 3D object detection works~\cite{ku2018joint, qi2017pointnet2, qi2018frustum}. 
In \cite{li2019gs3d, liu2019deep, manhardt2019roi, mousavian20173d, simonelli2019disentangling} strong 2D detection systems are extended to add cues such as object orientation, then the remaining 3D box parameters are solved via 3D box geometry.
\cite{brazil2019m3d} extends the region proposal network (RPN) of Faster R-CNN~\cite{ren2015faster} with 3D box parameters.

\SubSubSection{Orientation Estimation:}
Prior monocular 3D object detectors estimate orientation via two main techniques. 
The first method is to classify orientation via a series of discrete bins then regress a relative offset~\cite{chen20153d, chen2016monocular, ku2019monocular, li2019gs3d, liu2019deep, manhardt2019roi, mousavian20173d, xu2018multi}.
The bin technique requires a trade-off between the quantity/coverage of the discretized angles and an increase in the number of estimated parameters (bin $\times$).
Other methods directly regress the orientation angle using quaternion~\cite{simonelli2019disentangling} or Euler~\cite{brazil2019m3d, ma2019accurate} angles.
Direct regression is comparatively efficient, but may lead to degraded performance and periodicity challenges~\cite{zhou2019continuity}, as exemplified in Fig.~\ref{fig:teaser}.

In contrast, we propose a novel orientation decomposition which serves as an intuitive compromise between the bin and direct approaches. 
We decompose the orientation estimation into three components: axis and heading classification, followed by an angle offset. 
Thus, our technique increases the parameters by a \textit{static} factor of $2$ compared to a bin hyperparameter, while drastically reducing the offset search space for each orientation estimation (discussed in Sec.~\ref{sec:measurement}).

\SubSubSection{Uncertainty Estimation:}
Although it is common to utilize the classification score to rate boxes in 2D object detection~\cite{chabot2017deep, ren2015faster, redmon2016you, yang2016exploit} or explicitly model uncertainty as parametric estimation~\cite{luvli-face-alignment-estimating-landmarks-location-uncertainty-and-visibility-likelihood}, prior works in monocular 3D object detection realize the need for 3D box uncertainty/confidence~\cite{liu2019deep, simonelli2019disentangling}.
\cite{liu2019deep} defines confidence using the 3D IoU of a box and ground truth after center alignment, thus capturing the confidence primarily of the 3D \textit{object dimensions}.\linebreak
\cite{simonelli2019disentangling} predicts a confidence by re-mapping the 3D box loss into a probability range, which intuitively represents the confidence of the overall 3D box accuracy. 

In contrast, our \textit{self-balancing} confidence loss is generic and self-supervised, with two benefits. 
(1) It enables estimation of a 3D localization confidence using only the loss values, thus being more general than 3D IoU.
(2) It enables the network to naturally re-balance extremely hard 3D boxes and focus on relatively achievable samples. 
Our ablation (Sec.~\ref{sec:ablation}) shows the importance of both effects.

\SubSection{Video-based Object Detection}
\label{sec:related_video}
Video-based object detection~\cite{bertasius2018object, Liu_2018_CVPR, xiao2018video, zhu2017flow, zhu2018towards} is generally less studied than single-frame object detection~\cite{chabot2017deep, ren2015faster, redmon2016you, yang2016exploit, ren2017accurate, yang2019std}. 
A common trend in video-based detection is to benefit the accuracy-efficiency trade-off via reducing the frame redundancy~\cite{bertasius2018object, Liu_2018_CVPR, xiao2018video, zhu2017flow, zhu2018towards}.
Such works are applied primarily on domains of ImageNet VID 2015~\cite{russakovsky2015imagenet}, which contain less ego-motion from the capturing camera than self-driving scenarios~\cite{Geiger2012CVPR, cordts2016cityscapes}.
As such, the methods are designed to use 2D transformations, which lack the consistency and realism of 3D motion modeling. 

In comparison, to our knowledge this is the first work that \textit{utilizes video cues to improve the accuracy and robustness of monocular 3D object detection}.
In the domain of 2D/3D object tracking,~\cite{giancola2019leveraging} experiments using Kalman Filters, Particle Filters, and Gaussian Mixture Models, and observe Kalman to be the most effective aggregation method for tracking.
An LSTM with depth ordering and IMU camera ego-motion is utilized in~\cite{Hu_2019_ICCV} to improve the tracking accuracy. \linebreak
In contrast, we explore how to naturally and effectively leverage a 3D Kalman filter to improve the accuracy and robustness of monocular 3D object detection.
We propose novel enhancements including estimating ego-motion, orientation, and a 3D confidence, while efficiently using only a single model.  
\Section{Methodology}
Our proposed kinematic framework is composed of three primary components: \linebreak a 3D region proposal network (RPN), ego-motion estimation, and a novel kinematic model to \textit{take advantage of temporal motion in videos}. 
We first overview the foundations of a 3D RPN. 
Then we detail our contributions of orientation decomposition and self-balancing 3D confidence, which are integral to the kinematic method. 
Next we detail ego-motion estimation. 
Lastly, we present the complete kinematic framework (Fig.~\ref{fig:overview}) which carefully employs a 3D Kalman~\cite{welch1995introduction} to model realistic motion using the aforementioned components, ultimately producing a more accurate and comprehensive 3D scene understanding.

\begin{figure*}[t]
\begin{center}
   \includegraphics[trim = 0 20 0 0, clip,width=0.99\linewidth]{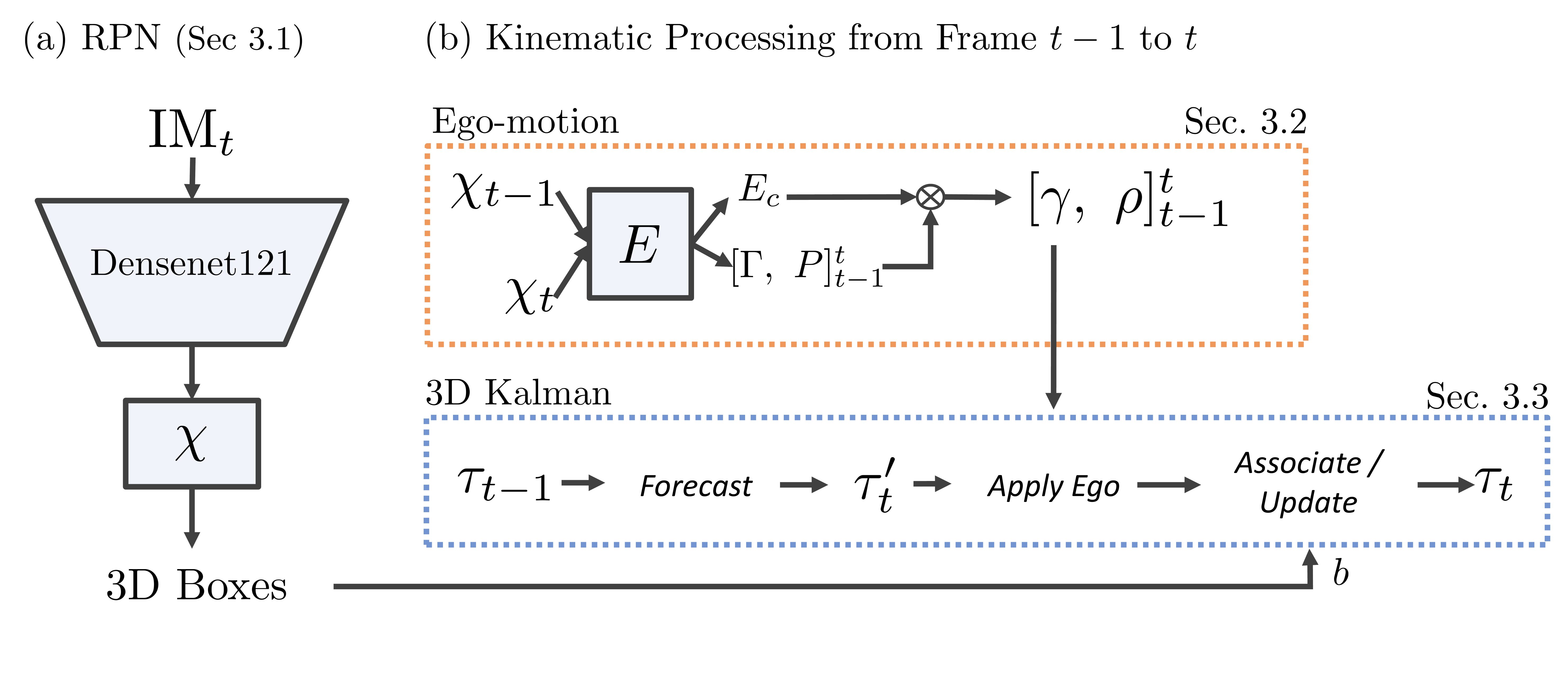}
\vspace{-5mm}
      \caption{
\textbf{Overview.} Our framework uses a RPN to first estimate 3D boxes (Sec.~\ref{sec:measurement}). 
We forecast previous frame tracks $\tau_{t-1}$ into $\tau'_t$ using the estimated Kalman velocity.
Self-motion is compensated for applying a global ego-motion (Sec.~\ref{sec:ego_motion}) to tracks $\tau'_{t}$. 
Lastly, we fuse $\tau'_t$ with measurements $b$ using a kinematic 3D Kalman filter (Sec.~\ref{sec:kalman}).
}
\label{fig:overview}
\end{center}\vspace{-6mm}
\end{figure*}

\SubSection{Region Proposal Network}
\label{sec:measurement}
Our measurement model is founded on the 3D RPN~\cite{brazil2019m3d}, enhanced using novel orientation and confidence estimations. 
The RPN itself acts as a sliding window detector following the typical practices outlined in Faster R-CNN~\cite{ren2015faster} and \cite{brazil2019m3d}. 
Specifically, the RPN consists of a backbone network and a detection head which predicts 3D box outputs relative to a set of predefined anchors. 

\SubSubSection{Anchors:}
We define our 2D-3D anchor $\Phi$ to consist of 2D dimensions $[\Phi_w, \Phi_h]_\two$ in pixels, a projected depth-buffer $\Phi_{z}$ in meters, 3D dimensions $[\Phi_w, \Phi_h, \Phi_l]_\three$ in meters, and orientations with respect to two major axes $[\Phi_{0}, \Phi_{1}]_\three$ in radians.
The term $\Phi_{z}$ is related to the camera coordinate $[x, y, z]_\three$ by the equation $\Phi_{z} \cdot [u, v, 1]_\two^\textmd{T} = \mathbf{\Upsilon} \cdot [x, y, z, 1]^\textmd{T}_\three$ where $\mathbf{\Upsilon} \in \mathbb{R}^{3\times4}$ is a known projection matrix.
We compute the anchor values by taking the mean of each parameter after clustering all ground truth objects in 2D, following the process in~\cite{brazil2019m3d}.

\SubSubSection{3D Box Outputs:}
Since our network is based on the principles of a RPN~\cite{brazil2019m3d, ren2015faster}, most of the estimations are defined as a transformation $\mathbf{T}$ relative to an anchor. \linebreak
Let us define $n_a$ as the number of anchors, $n_c$ as the number of object classes, and $w \times h$ as the output resolution of the network. 
The RPN outputs a classification map $\mathbf{C} \in \mathbb{R}^{(n_a \cdot n_c) \times w \times h}$, then 2D transformations $[\mathbf{T}_x, \mathbf{T}_y, \mathbf{T}_w, \mathbf{T}_h]_\two$, 3D transformations $[\mathbf{T}_u, \mathbf{T}_v, \mathbf{T}_z, \mathbf{T}_w, \mathbf{T}_h, \mathbf{T}_l, \mathbf{T}_{\theta_r}]_\three$, axis and heading $[\mathbf{\Theta}_a, \mathbf{\Theta}_h]$, and lastly a 3D self-balancing confidence $\mathbf{\Omega}$.
Each output has a size of $\mathbb{R}^{n_a \times w \times h}$.\linebreak
The outputs can be unrolled into $n_b = (n_a \cdot w \cdot h)$ boxes with $(n_c + 14)$-dim, with parameters of $c$, $[t_x, t_y, t_w, t_h]_\two$, $[t_u, t_v, t_z, t_w, t_h, t_l, t_{\theta_r}]_\three$, $[{\theta_a}, {\theta_h}]$, and $\omega$, \linebreak
which relate to the maps by $c\in \mathbf{C}$, $t_\two \in \mathbf{T}_\two$, $t_\three \in \mathbf{T}_\three$, $\theta \in \mathbf{\Theta}$ and $\omega \in \mathbf{\Omega}$.
The regression targets for 2D ground truths (GTs) $[\hat{x}, \hat{y}, \hat{w}, \hat{h}]_\two$ are defined as:
\vspace{-2mm}
\begin{alignat}{3}
\hat{t}_{x_\two} = \frac{\hat{x}_\two - i}{\Phi_{w_\two}},  ~&~ \hat{t}_{y_\two} = \frac{\hat{y}_\two - j}{\Phi_{h_\two}}, ~&~ \hat{t}_{w_\two} = \log{\frac{\hat{w}_\two}{\Phi_{w_\two}}}, ~&~ \hat{t}_{h_\two} = \log{\frac{\hat{h}_\two}{\Phi_{h_\two}}},
\vspace{-2mm}
\end{alignat}
where $(i, j) \in \mathbb{R}^{w \times h}$ represent the pixel coordinates of the corresponding box. 
Similarly, following the equation of $\hat{z}\cdot[\hat{u}, \hat{v}, 1]^\text{T}_\two = \mathbf{\Upsilon} \cdot [x, y, z, 1]^\text{T}_\three$, the regression targets for the projected 3D center GTs are defined as:
\vspace{-2mm}
\begin{alignat}{3}
\hat{t}_{u} = \frac{\hat{u} - i}{\Phi_{w_\two}},  ~~~~&~~~~ \hat{t}_{v} = \frac{\hat{v} - j}{\Phi_{h_\two}}, ~~~~&~~~~ \hat{t}_{z} = \hat{z} - \Phi_{z}.
\vspace{-2mm}
\end{alignat}
Lastly, the regression targets for 3D dimensions GTs $[\hat{w}, \hat{h}, \hat{l}]_\three$ are defined as:
\vspace{-2mm}
\begin{alignat}{3}
\hat{t}_{w_\three} = \log{\frac{\hat{w}_\three}{\Phi_{w_\three}}}, ~~~&~~~ \hat{t}_{h_\three} = \log{\frac{\hat{h}_\three}{\Phi_{h_\three}}}, ~~~&~~~ \hat{t}_{l_\three} = \log{\frac{\hat{l}_\three}{\Phi_{l_\three}}}.
\vspace{-2mm}
\end{alignat}
The remaining targets for our novel orientation estimation $t_{\theta_r}$, $[\theta_a,~\theta_h]$, and 3D self-balancing confidence $\omega$ are defined in subsequent sections. 

\begin{figure*}[t]
\begin{center}
   \includegraphics[trim = 0 10 0 0, clip, width=0.99\linewidth]{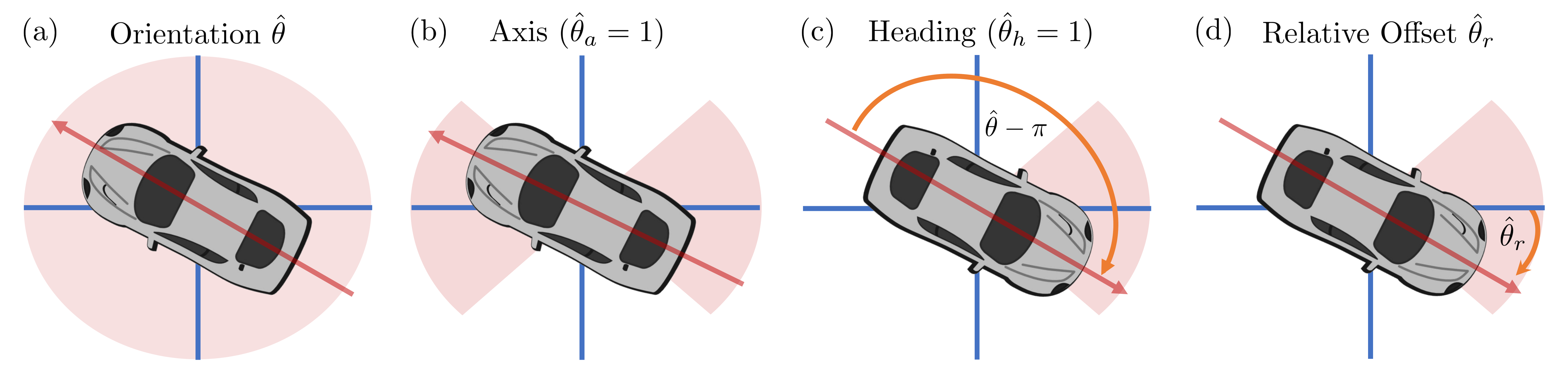}
\vspace{-1mm}
      \caption{
\textbf{Orientation.} Our proposed orientation formulation decomposes an object orientation $\hat{\theta}$ (a) into an axis classification $\hat{\theta}_a$ (b), a heading classification $\hat{\theta}_h$ (c), and an offset $\hat{\theta}_r$ (d). 
Our method disentangles the objectives of axis and heading classification while greatly reducing the offset region (red) by a factor of $\frac{1}{4}$.
}
\label{fig:orientation}
\end{center}\vspace{-6mm}
\end{figure*}

\SubSubSection{Orientation Estimation:}
We propose a novel object orientation formulation, with a decomposition of three components: axis, heading, and offset 
(Fig.~\ref{fig:orientation}). \linebreak
Intuitively, the axis estimation $\theta_a$ represents the probability an object is oriented towards the vertical axis ($\theta_a = 0$) or the horizontal axis ($\theta_a = 1$), with its label formally defined as:
$\hat{\theta}_a = \lvert \sin{\hat{\theta}} \rvert < \lvert \cos{\hat{\theta}} \rvert$,
where $\hat{\theta}$ is the ground truth object orientation in radians from a bird's eyes view (BEV) with $[-\pi, \pi)$ bounded range.\linebreak
We then compute an orientation $\hat{\theta}_r$ with a restricted range relative to its axis, \textit{e.g.}, $[-\pi,~0)$ when $\hat{\theta}_a = 0$, and $[-\frac{\pi}{2},~\frac{\pi}{2})$ when $\hat{\theta}_a = 1$.
We start with $\hat{\theta}_r = \hat{\theta}$ then add or subtract $\pi$ from $\hat{\theta}_r$ until the desired range is satisfied. 

Intuitively, $\hat{\theta}_r$ loses its heading since the true rotation may be $\{\hat{\theta}_r,~\hat{\theta}_r \pm \pi\}$.
We therefore preserve the heading using a separate $\hat{\theta}_h$, which represents the probability of $\hat{\theta}_r$ being rotated by $\pi$ with its GT target defined as:
\vspace{-2mm}
\begin{align}
\hat{\theta}_h = 
\begin{cases}
0 & \hat{\theta} = \hat{\theta}_r \\
1 & \textmd{otherwise}.
\end{cases}
\label{eqn:orientation_heading}
\vspace{-2mm}
\end{align}
Lastly, we encode the orientation offset transformation which is relative to the corresponding anchor, axis, and restricted orientation $\hat{\theta}_r$ as: $\hat{t}_{\theta_r} = {\hat{\theta}_r - \Phi_{\theta_a}}$. \linebreak
The reverse decomposition is $\theta = \Phi_{\theta_a} + \lfloor \theta_h \rceil \cdot \pi + t_{\theta_r}$ where $\lfloor \rceil$ denotes round.

In designing our orientation, we first observed that the visual difference between objects at opposite headings of $[\theta,~\theta \pm \pi]$ is low, especially for far objects. \linebreak
In contrast, classifying the axis of an object is intuitively more clear since the visual features correlate with the aspect ratio. 
Note that $[\theta_a,~\theta_h,~\theta_r]$ disentangle these two objectives. 
Hence, while the axis is being determined, the heading classifier can focus on subtle clues such as windshields, headlights and shape.

We further note that our $2$ binary classifications have the same representational power as $4$ bins following \cite{ku2019monocular, li2019gs3d, liu2019deep, manhardt2019roi}. 
Specifically, bins of $[0, \frac{\pi}{2}, \pi, \frac{3\pi}{2}]$. 
However, it is more common to use considerably more bins (such as $12$ in~\cite{ku2019monocular}). 
An important distinction is that bin-based approaches require the network decide axis and heading simultaneously, whereas our method \textit{disentangles} the orientation into the two distinct and explainable objectives. 
We provide ablations to compare our decomposition and the bin method using $[2, 4, 10]$ bins in Sec.~\ref{sec:ablation}.
 

\SubSubSection{Self-Balancing Loss:}
\label{sec:confidence}
The novel 3D localization confidence $\omega$ follows a self-balancing formulation closely coupled to the network loss. 
We first define the 2D and 3D loss terms which comprise the general RPN loss.
We unroll and match all $n_b$ box estimations to their respective ground truths.
A box is matched as foreground when sufficient ($\geq k$) 2D intersection over union (IoU) is met, otherwise it is considered background ($\hat{c} = 0$) and all loss terms except for classification are ignored. 
The 2D box loss is thus defined as:
\vspace{-2mm}
\begin{equation}
L_\two = -\log(\text{IoU}(b_\two,~\hat{b}_\two)) [\hat{c} \neq 0 ] + \text{CE}(c,~\hat{c}),
\vspace{-2mm}
\end{equation}
where CE denotes a softmax activation followed by logistic cross-entropy loss over the ground truth class $\hat{c}$, and IoU uses predicted $b_\two$ and ground truth $\hat{b}_\two$.
Similarly, the 3D localization loss for only foreground ($\hat{c} \neq 0$) is defined as:
\vspace{-2mm}
\begin{equation}
L_\three = L_1(t_\three,~\hat{t}_\three) + \lambda_a \cdot \textmd{BCE}([\theta_a,~\theta_h],~[\hat{\theta}_a,~\hat{\theta}_h]),
\vspace{-2mm}
\end{equation}
where BCE denotes a sigmoid activation followed by binary cross-entropy loss.
Next we define the final self-balancing confidence loss with the $\omega$ estimation as:
\vspace{-2mm}
\begin{equation}
L = L_\two + \omega \cdot L_\three + \lambda_L \cdot (1 - \omega),
\label{eqn:selfloss}
\vspace{-2mm}
\end{equation}
where $\lambda_L$ is the rolling mean of the $n_L$ most recent $L_\three$ losses per mini-batch.
Since $\omega$ is predicted per-box via a sigmoid, the network can intuitively balance whether to use the loss of $L_\three$ or incur a proportional penalty of $\lambda_L\cdot (1 - \omega)$. 
Hence, when the confidence is high ($\omega \approx 1$) we infer that the network is confident in its 3D loss $L_\three$.
Conversely, when the confidence is low ($\omega \approx 0$), the network is uncertain in $L_\three$, thus incurring a flat penalty is preferred. 
At inference, we fuse the self-balancing confidence with the classification score as $\mu = c \cdot \omega$.

The proposed self-balancing loss has two key benefits. 
Firstly, it produces a useful 3D localization confidence with inherent correlation to 3D IoU (Sec.~\ref{sec:ablation}).
Secondly, it enables the network to re-balance samples which are exceedingly challenging and re-focus on the more reasonable targets.
Such a characteristic can be seen as the inverse of hard-negative mining, which is important while  monocular 3D object detection remains highly difficult and unsaturated (Sec.~\ref{sec:exp_kitti}).

\SubSection{Ego-motion}
\label{sec:ego_motion}
A challenge with the dynamics of urban scenes is that not only are most foreground objects in motion, but the capturing camera itself is dynamic. 
Therefore, for a full understanding of the scene dynamics, we design our model to additionally predict the self-movement of the capturing camera, \textit{e.g.}, ego-motion.

We define ego-motion in the conventional six degrees of freedom: translation $[\gamma_x, \gamma_y, \gamma_z]$ in meters and rotation $[\rho_x, \rho_y, \rho_z]$ in radians. 
We attach an ego-motion features layer $E$ with ReLU to the concatenation of two temporally adjacent feature maps $\mathbf{\chi}$ which is the final layer in our backbone with size $\mathbb{R}^{n_h \times w \times h}$ architecture, defined as $\mathbf{\chi}_{t-1}\mid\mid\mathbf{\chi}_t$. 
We then attach predictions for translation $[\Gamma_x, \Gamma_y, \Gamma_z]$ and rotation $[P_x, P_y, P_z]$, 
which are of size $\mathbb{R}^{w \times h}$.
Instead of using a global pooling, we predict a spatial confidence map ${E}_c \in \mathbb{R}^{w \times h}$ based on $E$.\linebreak
We then apply softmax over the spatial dimension of $E_c$ such that $\sum{E_c} = 1$.
Hence, the pooling of prediction maps $[\Gamma,~P]$ into $[\gamma,~\rho]$ is defined as:
\begin{equation}
\gamma = \sum_{(i,j)} \Gamma (i, j) \cdot E_c(i, j),~~~~ \rho = \sum_{(i,j)} P (i, j) \cdot E_c(i, j),
\vspace{-2mm}
\end{equation}
where $(i, j)$ is the coordinate in $\mathbb{R}^{w \times h}$.
We show an overview of the motion features $E$, spatial confidence $E_c$, and outputs $[\Gamma,~P] \to [\gamma,~\rho]$ within Fig.~\ref{fig:overview}. \linebreak
We use a $L_1$ loss against GTs $\hat{\gamma}$ and $\hat{\rho}$ defined as
$L_\text{ego} = L_1(\gamma,~\hat{\gamma}) + \lambda_r \cdot L_1(\rho,~\hat{\rho})$.

\SubSection{Kinematics}
\label{sec:kalman}
In order to leverage temporal motion in video, we elect to integrate our RPN and ego-motion into a novel kinematic model. 
We adopt a 3D Kalman~\cite{welch1995introduction} due to its notable efficiency, effectiveness, and interpretability.
We next detail our proposed motion model, the procedure for forecasting, association, and update.

\SubSubSection{Motion Model:}
The critical variables we opt to track are defined as 3D center $[\tau_x, \tau_y, \tau_z]$, 3D dimensions $[\tau_w, \tau_h, \tau_l]$, orientation $[\tau_{\theta}, \tau_{\theta_h}]$, and scalar velocity $\tau_v$. \linebreak
We define $\tau_\theta$ as an $\theta$ orientation constrained to the range of $[-\frac{\pi}{2}, \frac{\pi}{2})$, and $\theta_h$ as in Sec.~\ref{sec:measurement}\footnote{We do not use the axis $\theta_a$ of Sec.~\ref{sec:measurement}, since we expect the orientation to change \textit{smoothly} and do not wish the orientation is \textit{relative} to a (potentially) changing axis.}. 
We constrain the motion model to only allow objects to move in the direction of their orientation. 
Hence, we define the state transition $\mathbf{F} \in \mathbb{R}^{9\times9}$ as:
\vspace{-2mm}
\begin{equation}
\mathbf{F} = 
\begin{bmatrix}
    \multirow{6}{*}{~\large{$\mathbf{I}^{9\times8}$}} &~~~~~\cos(\tau_\theta + \pi  \lfloor\tau_{\theta_h}\rceil)  \\
    &~~0 \\
    &~~-\sin(\tau_\theta + \pi  \lfloor\tau_{\theta_h}\rceil)  \\
    &~~ 0  \\[-0.75mm]
    &~~\vdots  \\
    &~~ 1  \\
  \end{bmatrix},
  \vspace{-2mm}
\end{equation}
where $\mathbf{I}$ denotes the identity matrix and the state variable order is respectively $[\tau_x, \tau_y, \tau_z, \tau_w, \tau_h, \tau_l, \tau_\theta, \tau_{\theta_h}, \tau_v]$.
We constrain the velocity to only move within its orientation to simplify the Kalman to work more effectively.
Recall that since our measurement model RPN processes a single frame, it does \textit{not} measure velocity.\linebreak
Thus, to map the tracked state space and measurement space, we also define an observation model $\mathbf{H}$ as a truncated identity map of size $\mathbf{I} \in \mathbb{R}^{8\times 9}$.\linebreak
We define covariance $\mathbf{P}$ with 3D confidence $\mu$, as $\mathbf{P} = \mathbf{I}^{9\times9} \cdot (1 - \mu) \cdot \lambda_o$
where $\lambda_o$ is an uncertainty weighting factor. 
Hence, we avoid the need to manually tune the covariance, while being dynamic to diverse and changing image content.

\SubSubSection{Forecasting:}
The forecasting step aims to utilize the tracked state variables and covariances of time $t-1$ to estimate the state of a future time $t$. 
The equation to forecast a state variable $\tau_{t-1}$ into $\tau'_t$ is:
$\tau_t' = \mathbf{F}_{t-1} \cdot \tau_{t-1}$,
where $\mathbf{F}_{t-1}$ is the state transition model at $t-1$. 
Note that both objects \textit{and} the capturing camera may have independent motion between consecutive frames. 
Therefore, we lastly apply the estimated ego-motion to all available tracks' 3D center $[\tau_x, \tau_y, \tau_z]$ by:
\vspace{-2mm}
\begin{equation}
\begin{bmatrix}
\mathbf{\tau}_x\\
\mathbf{\tau}_y\\
\mathbf{\tau}_z\\
1
\end{bmatrix}_t'
= 
\begin{bmatrix}
~\mathbf{R}, ~\mathbf{T}~\\
~0,~~~1~
\end{bmatrix}_{t-1}^t \cdot 
\begin{bmatrix}
\mathbf{\tau}_x\\
\mathbf{\tau}_y\\
\mathbf{\tau}_z\\
1
\end{bmatrix}_t',
~~~ \tau'_{t\theta} = \tau'_{t\theta} + \rho_y
\vspace{-2mm}
\end{equation}
where $\mathbf{R}_{t-1}^t \in \mathbb{R}^{3 \times 3}$ denotes the estimated rotation matrix converted from Euler angles and $\mathbf{T}_{t-1}^t \in \mathbb{R}^{3\times1}$ the translation vector for ego-motion (as in Sec.~\ref{sec:ego_motion}). 
Finally, we forecast a tracked object's covariance $\mathbf{P}$ from $t-1$ to $t$ defined as:
\vspace{-2mm}
\begin{equation}
\mathbf{P}_t' = \mathbf{F}_{t-1} \cdot \mathbf{P}_{t-1} \cdot \mathbf{F}_{t-1}^T + \mathbf{I}^{9\times9} \cdot (1 - \mu_{t-1}),
\vspace{-2mm}
\end{equation}
where $\mu_{t-1}$ denotes the \textit{average} self-balancing confidence $\mu$ of a track's life. 
Hence, the resultant track  states $\tau'_t$ and track covariances $\mathbf{P}'_t$ represent the Kalman filter's best forecasted estimation with respect to frame $t$.

\SubSubSection{Association:}
After the tracks have been forecasted from $t-1$ to $t$, the next step is to associate tracks to corresponding 3D box measurements (Sec.~\ref{sec:measurement}). 
Let us denote boxes produced by the measurement RPN as $b \in \mathbb{R}^8$ mimicking the tracked state as $[b_x, b_y, b_z, b_w, b_h, b_l, b_\theta, b_{\theta_h}]$\footnote{We apply the estimated transformations of Sec.~\ref{sec:measurement} to their respective anchors with equations of~\cite{brazil2019m3d}, and backproject into 3D coordinates to match track variables.}.
Our matching strategy consists of two steps.
We first compute the 3D center distance between the tracks $\tau_t'$ and measurements $b$.
The best matches with the lowest distance are iteratively paired and removed until no pairs remain with distance $\leq k_d$. 
Then we compute the \textit{projected} 2D box IoU between any remaining tracks $\tau_t'$ and measurements $b$.
The best matches with the highest IoU are also iteratively paired and removed until no pairs remain with IoU $\geq k_u$. 
Measured boxes that were \textit{not} matched are added as new tracks. 
Conversely, tracks that were \textit{not} matched incur a penalty with hyperparameter $k_p$, defined as $\mu_{t-1} = \mu_{t-1} \cdot k_p$. 
Lastly, any box who has confidence $\mu_{t-1} \leq k_m$ is removed from the valid tracks.

\SubSubSection{Update:}
After making associations between tracks $\tau_t'$ and measurements $b$, \linebreak the next step is to utilize the track covariance $\mathbf{P}_t'$ and measured confidence $\mu$ to update each track to its final state $\tau_t$ and covariance $\mathbf{P}_t$.
Firstly, we formally define the equation for computing the Kalman gain as:
\vspace{-2mm}
\begin{equation}
\mathbf{K} = \mathbf{P}' ~\mathbf{H}^T ~ (\mathbf{H}~\mathbf{P}'~\mathbf{H}^T + \mathbf{I}^{8\times8} ~ (1 - \mu)\cdot \lambda_o)^{-1},
\vspace{-2mm}
\end{equation}
where $\mathbf{I}^{8\times8} ~ (1 - \mu)\cdot \lambda_o$ represents the incoming measurement covariance matrix,
and $\mathbf{P}'$ the forecasted covariance of the track.
Next, given the Kalman gain $\mathbf{K}$, forecasted state $\tau'_t$, forecasted covariance $\mathbf{P}'_t$, and measured box $b$, the final track state $\tau_t$ and covariance $\mathbf{P}_t$ are defined as:
\vspace{-2mm}
\begin{equation}
\tau_t = \tau'_t + \mathbf{K} ~ (b - \mathbf{H}~\tau'_t), \hspace{8mm} 
\mathbf{P}_t = (\mathbf{I}^{9\times9} - \mathbf{K} ~ \mathbf{H})~\mathbf{P}_t'.
\vspace{-2mm}
\end{equation}
We lastly aggregate each track's overall confidence $\mu_{t}$ over time as a running average of
$\mu_t = \small{\frac{1}{2}} \cdot (\mu_{t-1} + \mu)$,
where $\mu$ is the measured confidence. 

\SubSection{Implementation Details}
\label{sec:implementation}
Our framework is implemented in PyTorch~\cite{paszke2017automatic}, with the 3D RPN settings of~\cite{brazil2019m3d}.\linebreak 
We release source code at \small\url{http://cvlab.cse.msu.edu/project-kinematic.html}\normalsize. 
We use a batch size of $2$ and learning rate of $0.004$.
We set $k=0.5$, $\lambda_o = 0.2$, $\lambda_r=40$, $n_L=100$, $\lambda_a=k_u=0.35$, $k_d=0.5$, $k_p=0.75$, and $k_m=0.05$.\linebreak
To ease training, we implement three phases. 
We first train the 2D-3D RPN with $L = L_\two + L_\three$, then the self-balancing loss of Eq.~\ref{eqn:selfloss}, for $80k$ and $50k$ iterations.
We freeze the RPN to train ego-motion using $L_\text{ego}$ for $80k$.
Our backbone is DenseNet121~\cite{huang2017densely} where $n_h=1{,}024$.
Inference uses $4$ frames as provided by~\cite{Geiger2012CVPR}. 

\definecolor{slightgray}{rgb}{0.35, 0.35, 0.35}
\definecolor{darkgreen}{rgb}{0.0, 0.5, 0.0}

\Section{Experiments}

We benchmark our kinematic framework on the KITTI~\cite{Geiger2012CVPR} dataset.
We comprehensively evaluate on 3D Object Detection and Bird's Eye View (BEV) tasks.\linebreak
We then provide ablation experiments to better understand the effects and justification of our core methodology. 
We show qualitative examples in Fig.~\ref{fig:qual}. 

\SubSection{KITTI Dataset}
\label{sec:exp_kitti}
The KITTI~\cite{Geiger2012CVPR} dataset is a popular benchmark for self-driving tasks. 
The official dataset consists of $7{,}481$ training and $7{,}518$ testing images including annotations for 2D/3D objects, ego-motion, and $4$ temporally adjacent frames.
We evaluate on the most widely used validation split as proposed in~\cite{chen20153d}, which consists of $3{,}712$ training and $3{,}769$ validation images. 
We focus primarily on the car class. 

\SubSubSection{Metric:}
Average precision (AP) is utilized for object detection in KITTI. \linebreak
Following~\cite{simonelli2019disentangling}, the KITTI metric has updated to include $40$ ($\uparrow$ $11$) recall points while \textit{skipping} the first. 
The AP$_{40}$ metric is more stable and fair overall~\cite{simonelli2019disentangling}.
Due to the official adoption of AP$_{40}$, it is not possible to compute AP$_{11}$ on test.
Hence, we elect to use the AP$_{40}$ metric for all reported experiments.
\begin{table*}[t!]
\begin{center}
\setlength\tabcolsep{10.00pt}
  \resizebox{\textwidth}{!}{  
\begin{tabular}{l | c c c  | c c c | c}
\hline
\multirow{2}{*}{ } & \multicolumn{3}{c|}{AP$_\three$ \small{(IoU $\geq 0.7$)}} & \multicolumn{3}{c|}{AP$_\text{BEV}$ \small{(IoU $\geq 0.7$)}} & \multirow{2}{*}{s/im$^*$ }  \\ 
& Easy & Mod & Hard & Easy & Mod & Hard \\ 
\hline
FQNet~\cite{liu2019deep} & ~~$2.77$ & ~~$1.51$ & ~~$1.01$ & ~~$5.40$ & ~~$3.23$ & ~~$2.46$ & $0.50^\dagger$ \\
ROI-10D~\cite{manhardt2019roi} & ~~$4.32$ & ~~$2.02$ & ~~$1.46$ & ~~$9.78$ & ~~$4.91$ & ~~$3.74$ & $0.20$~ \\
GS3D~\cite{li2019gs3d} & ~~$4.47$ & ~~$2.90$ & ~~$2.47$ & ~~$8.41$ & ~~$6.08$ & ~~$4.94$ & $2.00^\dagger$ \\
MonoPSR~\cite{ku2019monocular} & $10.76$ & ~~$7.25$ & ~~$5.85$ & $18.33$ & $12.58$ & ~~$9.91$ & $0.20$~ \\
MonoDIS~\cite{simonelli2019disentangling} & $10.37$ & ~~$7.94$ & ~~$6.40$ & $17.23$ & $13.19$ & $11.12$ & $-$~ \\
M3D-RPN~\cite{brazil2019m3d} & $14.76$ & ~~$9.71$ & ~~$7.42$ & $21.02$ & $13.67$ & $10.23$ & $\mathit{0.16}$~ \\
AM3D~\cite{ma2019accurate} & $\mathit{16.50}$ & $\mathit{10.74}$ & ~~$\mathbf{9.52}$ & $\mathit{25.03}$ & $\mathit{17.32}$ & $\mathbf{14.91}$ & $0.40$~ \\
\hline\hline
Ours & $\mathbf{19.07}$ & $\mathbf{12.72}$ & ~~$\mathit{9.17}$ & $\mathbf{26.69}$ & $\mathbf{17.52}$ & $\mathit{13.10}$ & $\mathbf{0.12}$~ \\
\hline
\end{tabular}
}
\end{center}
\caption{
\textbf{KITTI Test}. We compare with SOTA methods on the KITTI test dataset. 
We report performances using the AP$_{40}$~\cite{simonelli2019disentangling} metric available on the official leaderboard. 
*~the runtime is reported from the official leaderboard with slight variances in hardware. 
We indicate methods reported on CPU with $\dagger$. 
\textbf{Bold}/\textit{italics} indicate best/second AP.
}\label{tab:kitti_test}
\vspace{-2mm}
\end{table*}

\begin{table*}[t!]
\begin{center}
\setlength\tabcolsep{3.00pt}
  \resizebox{\textwidth}{!}{  
\begin{tabular}{l | c c c  | c c c }
\hline
\multirow{2}{*}{ } & \multicolumn{3}{c|}{AP$_\three$ \small{(IoU $\geq [0.7/0.5]$)}} & \multicolumn{3}{c}{AP$_\text{BEV}$ \small{(IoU $\geq [0.7/0.5]$)}}  \\ 
& Easy & Mod & Hard & Easy & Mod & Hard \\ 
\hline
MonoDIS~\cite{simonelli2019disentangling} & $11.06$/~~$-$~~ & ~~$7.60$/~~$-$~~ & ~~$6.37$/~~$-$~~  & $18.45$/~~$-$~~ & $12.58$/~~$-$~~ & $10.66$/~~$-$~~ \\
M3D-RPN~\cite{brazil2019m3d} & $\mathit{14.53}$/$\mathit{48.56}$ & $\mathit{11.07}$/$\mathit{35.94}$ & ~~$\mathit{8.65}$/$\mathit{28.59}$  & $\mathit{20.85}$/$\mathit{53.35}$ & $\mathit{15.62}$/$\mathit{39.60}$ & $\mathit{11.88}$/$\mathit{31.77}$\\
\hline\hline
Ours & $\mathbf{19.76}$/$\mathbf{55.44}$ & $\mathbf{14.10}$/$\mathbf{39.47}$ & $\mathbf{10.47}$/$\mathbf{31.26}$  & $\mathbf{27.83}$/$\mathbf{61.79}$ & $\mathbf{19.72}$/$\mathbf{44.68}$ & $\mathbf{15.10}$/$\mathbf{34.56}$\\
\hline
\end{tabular}
}
\end{center}
\caption{
\textbf{KITTI Validation}. We compare with SOTA on KITTI validation~\cite{chen20153d} split. 
Note that methods published prior to~\cite{simonelli2019disentangling} are unable to report the AP$_{40}$ metric.
}\label{tab:kitti_val}
\vspace{-6mm}
\end{table*}
\SubSection{3D Object Detection}
We evaluate our proposed framework on the task of 3D object detection, which requires objects be localized in 3D camera coordinates as well as supplying the 3D dimensions and BEV orientation relative to the XZ plane.
Due to the strict requirements of IoU in three dimensions, the task demands precise localization of an object to be considered a match (3D IoU $\geq 0.7$).
We evaluate our performance on the official test~\cite{Geiger2012CVPR} dataset in Tab.~\ref{tab:kitti_test} and the validation~\cite{chen20153d} split in Tab.~\ref{tab:kitti_val}. 

We emphasize that our method improves the SOTA on KITTI test by a significant margin of $\uparrow 1.98\%$ compared to~\cite{ma2019accurate} on the moderate configuration with IoU $\geq 0.7$, which is the most common metric used to compare.
Further, we note that \cite{ma2019accurate} require multiple encoder-decoder networks which add overhead compared to our single network approach.
Hence, their runtime is $\approx 3\times$ (Tab.~\ref{tab:kitti_test}) compared to ours, self-reported on \textit{similar} but not identical GPU hardware.
Moreover,~\cite{brazil2019m3d} is the most comparable method to ours as both utilize a single network and an RPN archetype. 
We note that our method significantly outperforms~\cite{brazil2019m3d} and many other recent works~\cite{ku2019monocular, li2019gs3d, liu2019deep, manhardt2019roi, simonelli2019disentangling } by $\approx 3.01-11.21\%$.

We further evaluate our approach on the KITTI validation~\cite{chen20153d} split using the AP$_{40}$ for available approaches and observe similar overall trends as in Tab.~\ref{tab:kitti_val}. \linebreak
For instance, compared to competitive approaches~\cite{brazil2019m3d, simonelli2019disentangling} our method improves the performance by $\uparrow 3.03\%$ for the challenging IoU criteria of $\geq 0.7$.
Similarly, our performance on the more relaxed criteria of IoU $\geq 0.5$ increases by $\uparrow 3.53\%$. \linebreak
We additionally visualize detailed performance characteristics on AP$_\three$ at discrete depth $[15,~30,~\text{All}]$ meters and IoU matching criterias $0.3\to0.7$ in Fig.~\ref{fig:ROC}.

\SubSection{Bird's Eye View}
The Bird's Eye View (BEV) task is similar to 3D object detection, differing primarily in that the 3D boxes are firstly projected into the XZ plane then 2D object detection is calculated. 
The projection collapses the Y-axis degree of freedom and intuitively results in a less precise but reasonable localization. 

We note that our method achieves SOTA performance on the BEV task regarding the moderate setting of the KITTI test dataset as detailed in Tab.~\ref{tab:kitti_test}. \linebreak
Our method performs favorably compared with SOTA works \cite{brazil2019m3d, ku2019monocular, li2019gs3d, liu2019deep, manhardt2019roi, simonelli2019disentangling} (\textit{e.g.}, $\approx 3.85 - 14.29\%$), and similarly to\cite{ma2019accurate} at a notably lower runtime cost. 
We suspect that our method, especially the self-balancing confidence (Eq.~\ref{eqn:selfloss}), prioritizes precise localization which warrants more benefit in \textbf{full} 3D Object Detection task compared to the Bird's Eye View task. 

Our method performs similarly on the validation~\cite{chen20153d} split of KITTI (Tab.~\ref{tab:kitti_val}).
Specifically, compared to \cite{brazil2019m3d, simonelli2019disentangling} our proposed method outperforms by a range of $\approx 4.10 - 7.14\%$, which is \textbf{consistent} to the same methods on test $\approx 3.85 - 4.33\%$.

\begin{table*}[t!]
\begin{center}
\setlength\tabcolsep{3.00pt}
  \resizebox{\textwidth}{!}{  
\begin{tabular}{l | c c c | c c c }
\hline
\multirow{2}{*}{ } & \multicolumn{3}{c|}{AP$_\three$ \small{(IoU $\geq [0.7/0.5]$)}} & \multicolumn{3}{c}{AP$_\text{BEV}$ \small{(IoU $\geq [0.7/0.5]$)}}  \\ 
& Easy & Mod & Hard & Easy & Mod & Hard \\ 
\hline
Baseline & $13.81$/$47.10$ & ~~$9.71$/$34.14$ & ~~$7.44$/$26.90$ & $20.08$/$52.57$ & $13.98$/$38.45$ & ~~$11.10$/$29.88$\\
+ $\theta$ decomposition & $16.66$/$51.47$ & $12.10$/$38.58$ & ~~$9.40$/$30.98$ & $23.15$/$56.48$ & $17.43$/$42.53$ & $13.48$/$34.37$\\
+ self-confidence & $16.64$/$52.18$ & $12.77$/${38.99}$ & ~~${9.60}$/$\mathbf{31.42}$ & $24.22$/$58.52$ & $18.02$/$42.95$ & $13.92$/$\mathbf{34.80}$\\
+ $\mu = c \cdot \omega$ & $\mathit{18.28}$/$\mathit{54.70}$ & $\mathit{13.55}$/$\mathit{39.33}$ & $\mathit{10.13}$/${31.25}$ & $\mathit{25.72}$/$\mathit{60.87}$ & $\mathit{18.82}$/$\mathit{44.36}$ & $\mathit{14.48}$/$34.48$\\
+ kinematics & $\mathbf{19.76}$/$\mathbf{55.44}$ & $\mathbf{14.10}$/$\mathbf{39.47}$ & $\mathbf{10.47}$/$\mathit{31.26}$  & $\mathbf{27.83}$/$\mathbf{61.79}$ & $\mathbf{19.72}$/$\mathbf{44.68}$ & $\mathbf{15.10}$/$\mathit{34.56}$ \\
\hline
\end{tabular}
}
\end{center}
\caption{
\textbf{Ablation Experiments}. We conduct a series of ablation experiments with the validation~\cite{chen20153d} split of KITTI, using diverse IoU matching criteria of $\geq 0.7/0.5$. 
}\label{tab:kitti_ablation}
\vspace{-8mm}
\end{table*}
\SubSection{Ablation Study}
\label{sec:ablation}

To better understand the characteristics of our proposed kinematic framework, we perform a series of ablation experiments and analysis, summarized in Tab.~\ref{tab:kitti_ablation}.
We adopt~\cite{brazil2019m3d} without hill-climbing or depth-aware layers as our baseline method.
Unless otherwise specified we use the experimental settings outlined in Sec.~\ref{sec:implementation}.

\SubSubSection{Orientation Improvement:}
The orientation of objects is intuitively a critical component when modeling motion. 
When the orientation is decomposed into axis, heading, and offset the overall performance significantly improves, \textit{e.g.}, by $\uparrow 2.39\%$ in AP$_\three$ and $\uparrow 3.45\%$ in AP$_\text{BEV}$, as detailed within Tab.~\ref{tab:kitti_ablation}. \linebreak
We compute the mean angle error of our baseline, orientation decomposition, and kinematics method which respectively achieve $13.4^{\circ}$, $10.9^{\circ}$, and $6.1^{\circ}$ ($\downarrow 54.48\%$), suggesting our proposed methodology is significantly more \textit{stable}. 

We compare our orientation decomposition to bin-based methods following general idea of~\cite{ku2019monocular, li2019gs3d, liu2019deep, manhardt2019roi}.
We specifically change our orientation definition into $[\theta_b,~\theta_o]$ which includes a bin classification and an offset. 
We experiment with the number of bins set to $[2,~4,~10]$ which are uniformly spread from $[0,~2\pi)$. \linebreak
Note that $4$ bins have the same representational power as using binary $[\theta_a,~\theta_h]$. 
We observe that the ablated bin-based methods achieve $[9.47\%,~10.02\%,~10.76\%]$ in AP$_\three$. 
In comparison, our decomposed orientation achieves $12.10\%$ in AP$_\three$. 
We provide additional detailed experiments in our supplemental material.

Further, we find that our proposed kinematic motion model (as in Sec.~\ref{sec:kalman}) \textit{degrades} in performance when a comparatively erratic baseline (Row 1. Tab.~\ref{tab:kitti_ablation}) orientation is utilized instead ($14.10 \to 11.47$ on AP$_\three$), reaffirming the importance of having a consistent/stable orientation when aggregating through time.

\SubSubSection{Self-balancing Confidence:}
We observe that the self-balancing confidence is important from two key respects. 
Firstly, its integration in Eq.~\ref{eqn:selfloss} enables the network to re-weight box samples to focus more on reasonable samples and incur a flat penalty (\textit{e.g.}, $\lambda_L \cdot (1 - \omega)$ of Eq.~\ref{eqn:selfloss}) on the difficult samples. \linebreak
In a sense, the self-balancing confidence loss is the \textit{inverse} of hard-negative mining, allowing the network to focus on reasonable estimations.
Hence, the loss on its own improves performance for AP$_\three$ by $\uparrow 0.67\%$ and AP$_\text{BEV}$ by $\uparrow 0.59\%$. 

The second benefit of self-balancing confidence is that by design $\omega$ has an inherent correlation with the 3D object detection performance. 
Recall that we fuse $\omega$  with the classification score $c$ to produce a final box rating of $\mu = c~\cdot~\omega$, which results in an additional gain of $\uparrow 0.78\%$ in AP$_\three$ and $\uparrow 0.80\%$ in AP$_\text{BEV}$.
We further analyze the correlation of $\mu$ with 3D IoU, as is summarized in Fig.~\ref{fig:correlation}. 
The correlation coefficient with the classification score $c$ is significantly lower than the correlation using $\mu$ instead ($0.301$ vs. $0.417$).
In summary, the use of the Eq.~\ref{eqn:selfloss} and $\mu$ account for a gain of $\uparrow 1.45\%$ in AP$_\three$ and $\uparrow 1.39\%$ in AP$_\text{BEV}$.

\begin{figure} [t]
\centering
\begin{minipage}{0.47\linewidth}
   \includegraphics[width=\linewidth]{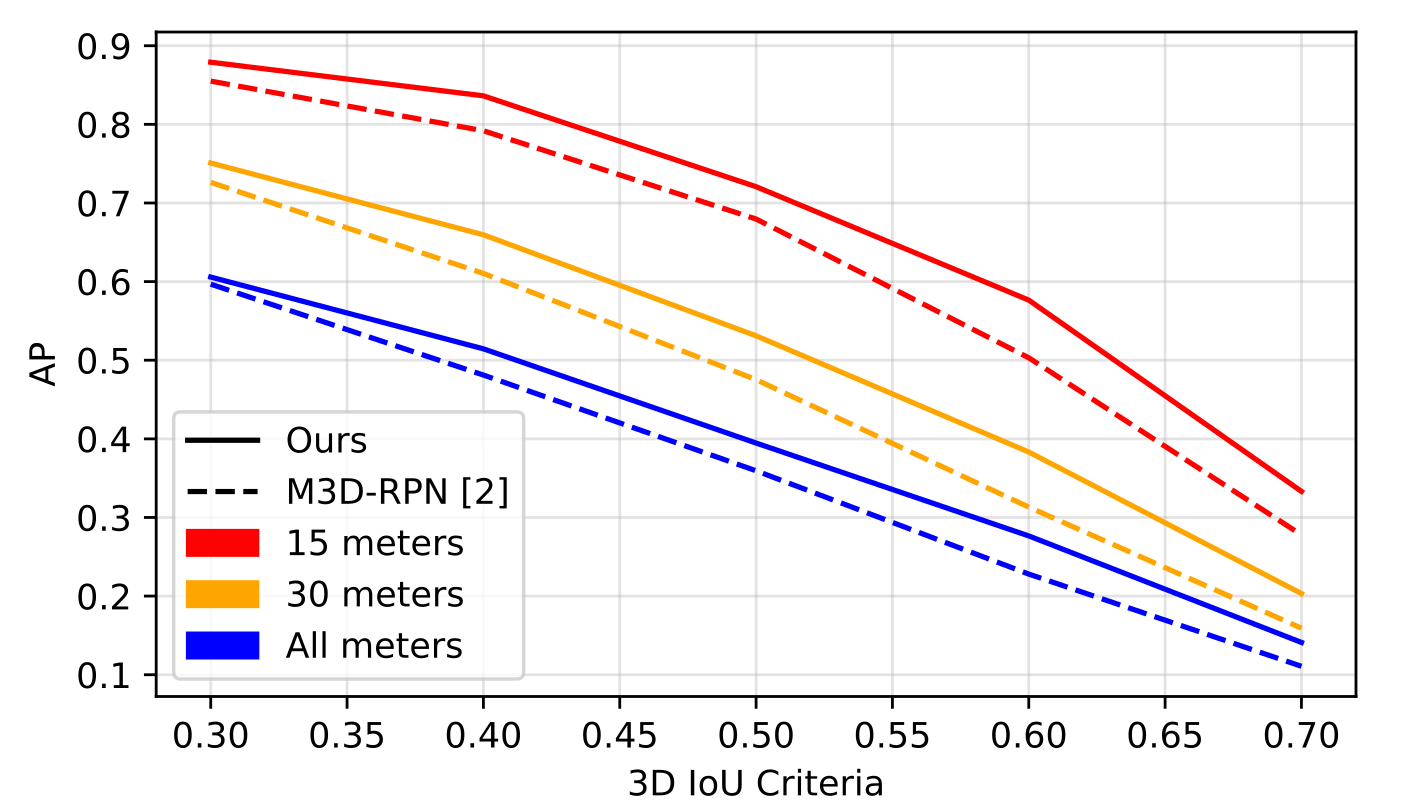}
\vspace{-7mm}
\caption{
We compare AP$_\three$ with \cite{brazil2019m3d} by varying 3D IoU criteria \textit{and} depth.
}
\label{fig:ROC}
\end{minipage}%
    \hfill%
\begin{minipage}{0.49\linewidth}
   \includegraphics[width=0.99\linewidth]{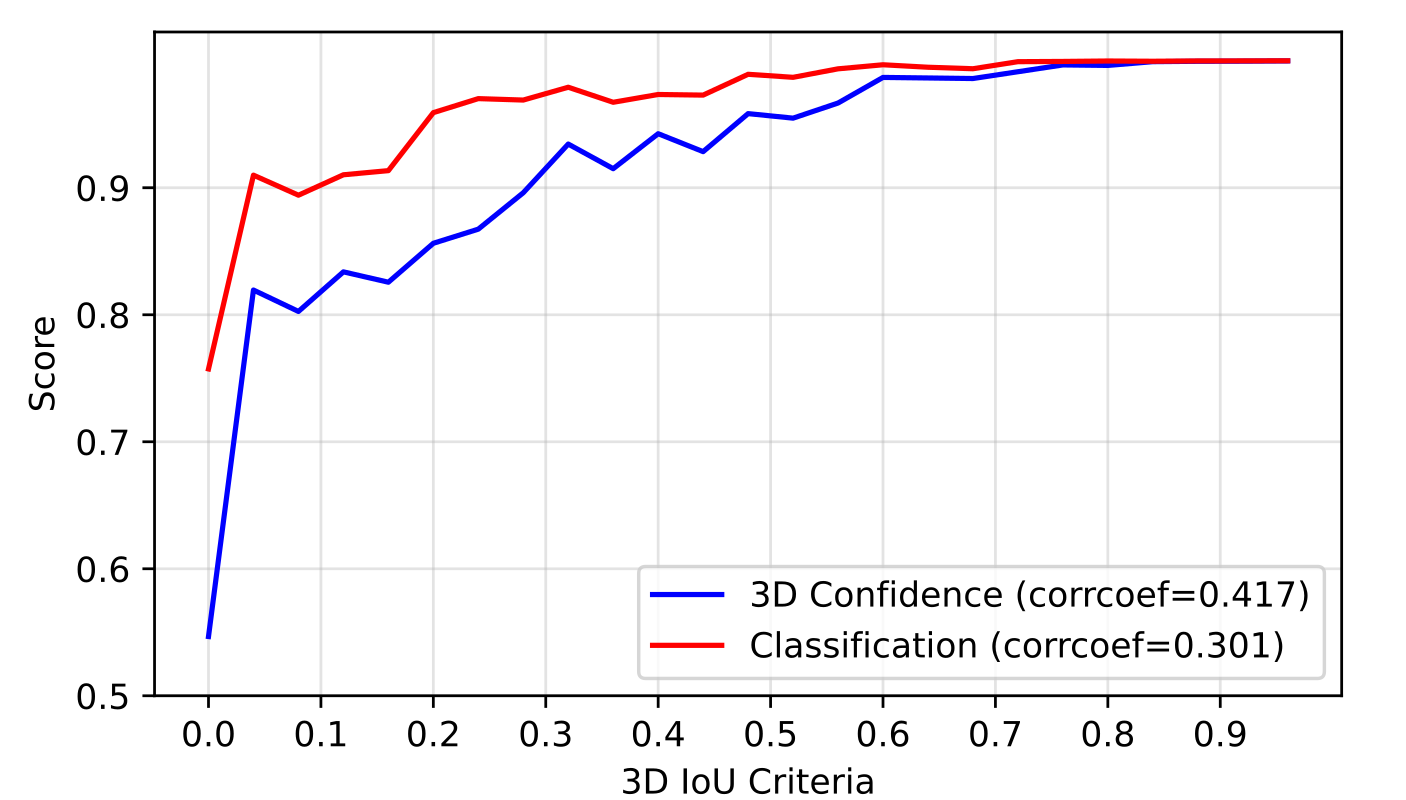}
\vspace{-7mm}
\caption{
We show the correlation of 3D IoU to classification $c$ and 3D confidence $\mu$. }
\label{fig:correlation}
\end{minipage} \vspace{-6mm}
\end{figure}

\SubSubSection{Temporal Modeling:}
The use of video and kinematics is a significant motivating factor for this work.
We find that the use of kinematics (detailed in Sec.~\ref{sec:kalman}) \linebreak results in a gain of $\uparrow 0.55\%$ in AP$_\three$ and $\uparrow 0.90\%$ in AP$_\text{BEV}$, as shown in Tab.~\ref{tab:kitti_ablation}.
We emphasize that although the improvement is less dramatic versus orientation and self-confidence, the former are important to facilitate temporal modeling. \linebreak
We find that if orientation decomposition and uncertainty are \textbf{removed}, by using the baseline orientation and setting $\mu$ to be a static constant, then the kinematic performance drastically reduces from $14.10\% \to 10.64\%$ in AP$_\three$.


We emphasize that kinematic framework not only helps 3D object detection, but also \textit{naturally} produces useful by-products such as velocity and ego-motion.
Thus, we evaluate the respective average errors of each motion after applying the camera capture rate to convert the motion into miles per hour (MPH). \linebreak
We find that the per-object velocity and ego-motion speed errors perform reasonably at $7.036$ MPH and $6.482$ MPH respectively.
We depict visual examples of all dynamic by-products in Fig.~\ref{fig:qual} and additionally in supplemental video.
\begin{figure*}[t]
\vspace{-4mm}
\begin{center}
   \includegraphics[trim = 0 0 0 0, clip, width=0.99\linewidth]{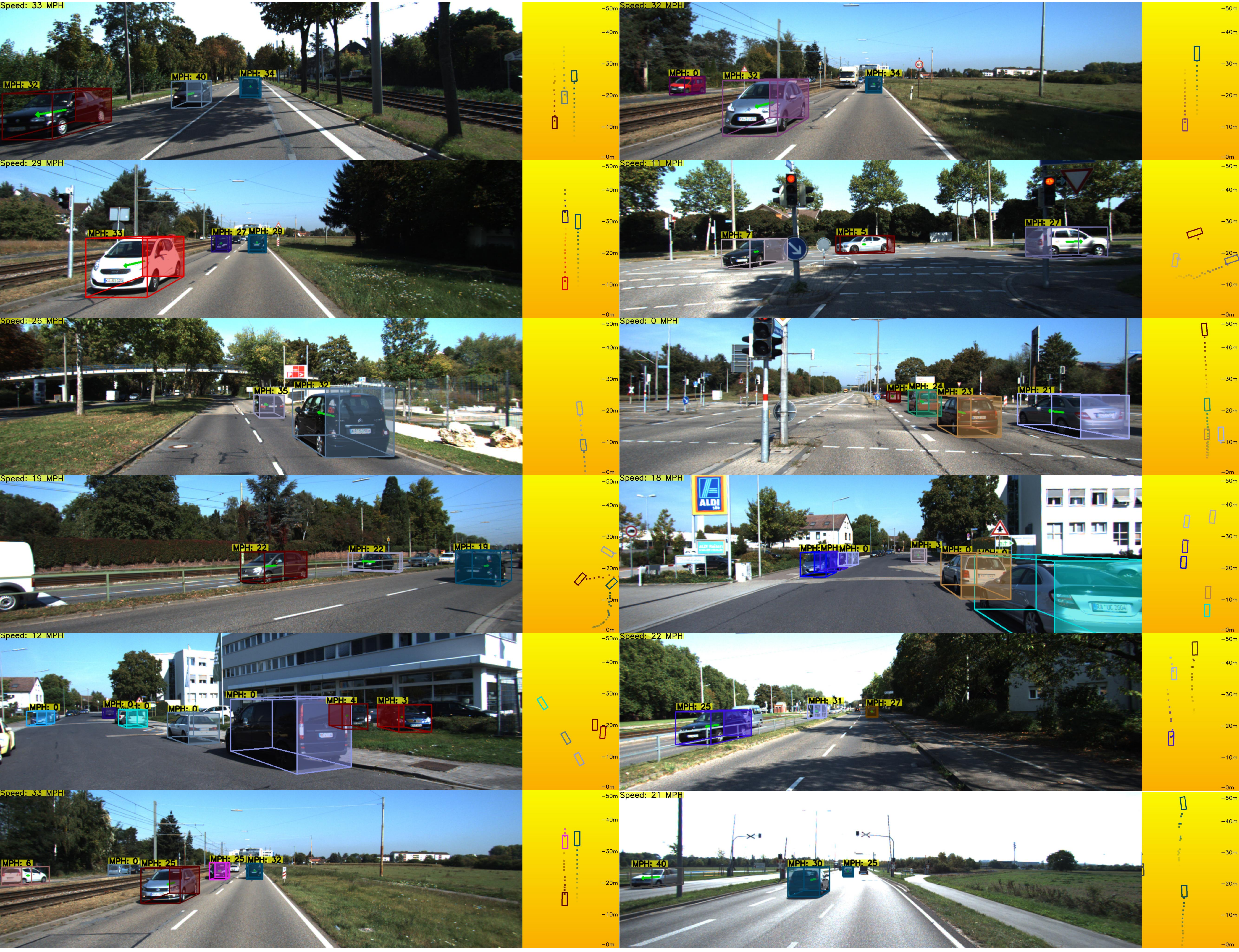}
\vspace{-2mm}
      \caption{
\textbf{Qualitative Examples.} 
We depict the image view (left) and BEV (right). 
We show velocity vector in green, speed and ego-motion in miles per hour (MPH) on top of detection boxes and at the top-left corner, and tracks as dots in BEV. 
}
\label{fig:qual}
\end{center}\vspace{-7mm}
\end{figure*}
\vspace{-3mm}
\Section{Conclusions}
We present a novel kinematic 3D object detection framework which is able to efficiently leverage temporal cues and constraints to improve 3D object detection. 
Our method naturally provides useful by-products regarding scene dynamics, \textit{e.g.}, reasonably accurate ego-motion and per-object velocity. 
We further propose novel designs of orientation estimation and a self-balancing 3D confidence loss in order to enable the proposed kinematic model to work effectively. 
We emphasize that our framework efficiently uses only a \textit{single network} to comprehensively understand a highly dynamic 3D scene for urban autonomous driving. 
Moreover, we demonstrate our method's effectiveness through detailed experiments on the KITTI~\cite{Geiger2012CVPR} dataset across the 3D object detection and BEV tasks.

\Paragraph{Acknowledgments:}
Research was partially sponsored by the Army Research Office under Grant Number W911NF-18-1-0330. The views and conclusions contained in this
document are those of the authors and should not be interpreted as representing the official policies, either expressed or implied, of the Army Research Office or the U.S. Government. The U.S. Government is authorized to reproduce and distribute reprints for Government purposes notwithstanding any copyright notation herein. 
This work is further partly funded by the Deutsche Forschungsgemeinschaft (DFG, German Research Foundation) - 409792180 (Emmy Noether Programme, project: Real Virtual Humans).

\clearpage

\setcounter{equation}{0}
\setcounter{figure}{0}
\setcounter{table}{0}
\setcounter{section}{0}

\title{Supplementary Material: Kinematic 3D Object Detection in Monocular Video}

\titlerunning{Kinematic 3D Object Detection in Monocular Video}
%
\author{Garrick Brazil\inst{1} \and
Gerard Pons-Moll\inst{2} \and
Xiaoming Liu\inst{1} \and Bernt Schiele\inst{2}}
\authorrunning{G. Brazil et al.}
%
\institute{Michigan State University, Computer Science $\&$ Engineering \\ \and
Max Planck Institute for Informatics, Saarland Informatics Campus\\
\inst{1}~\email{\{brazilga,~liuxm\}@msu.edu},~~\inst{2}~\email{\{gpons,~schiele\}@mpi-inf.mpg.de}
}

\maketitle

\begin{table*}
\begin{center}
\setlength\tabcolsep{2.00pt}
  \resizebox{\textwidth}{!}{  
\begin{tabular}{c | c c c | c c c }
\hline
\multirow{2}{*}{ } & \multicolumn{3}{c|}{AP$_\three$ \small{(IoU $\geq [0.7/0.5]$)}} & \multicolumn{3}{c}{AP$_\text{BEV}$ \small{(IoU $\geq [0.7/0.5]$)}}  \\ 
& Easy & Mod & Hard & Easy & Mod & Hard \\ 
\hline
$2$ bins & $12.83$/$46.46$ & ~~$9.47$/$33.78$ & ~~$7.93$/$26.85$ & $19.17$/$52.06$ & $14.72$/$37.54$ & $11.38$/$31.16$\\
$4$ bins & $12.65$/$44.01$ & $10.02$/${33.27}$ & ~~${7.87}$/${26.27}$ & $19.09$/$49.86$ & $14.55$/$37.90$ & $11.14$/${30.44}$\\
$10$ bins & ${14.27}$/${49.71}$ & ${10.74}$/${36.12}$ & ~~${8.29}$/${28.62}$ & ${21.12}$/${54.70}$ & ${15.37}$/${39.72}$ & ${11.60}$/$31.75$\\
\hline
Our decomp. & $16.66$/$51.47$ & $12.10$/$38.58$ & ~~$9.40$/$30.98$ & $23.15$/$56.48$ & $17.43$/$42.53$ & $13.48$/$34.37$\\
\hline
\end{tabular}
}
\end{center}
\caption{
\textbf{Orientation}.
We compare our orientation decomposition to bin-based orientation following the high-level concepts within~\cite{ku2019monocular, li2019gs3d, liu2019deep, manhardt2019roi}, using AP$_\three$ and AP$_\text{BEV}$. 
We evaluate our performances on the KITTI validation set~\cite{chen20153d} using IoU$~\geq~0.7/0.5$.
} \label{tab:orientation}
\vspace{-8mm}
\end{table*}

\section{Orientation Ablations}
We provide detailed experiments on 3D object detection and Bird's Eye View tasks to compare our orientation decomposition performance with bin-based approaches such as~\cite{ku2019monocular, li2019gs3d, liu2019deep, manhardt2019roi} within Tab.~\ref{tab:orientation}. 
Recall that bin-based orientation first classifies the best bin for orientation then predicts an offset with respect to the bin. 
In contrast, our method disentangles the bin classification into a distinct explainable objectives such as an axis classification and a heading classification. 
For such experiments we change our formulation to use bins of $[2,~4,~10]$, where $4$ bins has a similar representational power as two binary classifications $[\theta_a,~\theta_h]$. \linebreak
The bins are spread uniformly from $[0,~2\pi]$ and an offset is predicted afterwards.
We use the settings in Sec.~$3.4$ in main paper.
We emphasize that our method outperforms the bin-based approaches between $\approx 1.36 - 2.63\%$ on AP$_\three$ and $\approx 2.06 - 2.71\%$ on AP$_\text{BEV}$ using the standard moderate setting and IoU$~\geq~0.7$.

\section{Kalman Forecasting}

Since our method uses ego-motion and a 3D Kalman filter to aggregate temporal information, the approach can be modified to act as a box forecaster.
Although our method was not strictly designed for the tracking and forecasting task, we evaluate the 3D object detection and Bird's Eye View performance after forecasting $n_f=[1,~2,~3,~4]$ frames into the future.  
We assume a static ego-motion for unknown frames and otherwise use the Kalman equations described in the main paper Sec. 3.3 to forecast the tracked boxes. 

For all forecasting experiments we process $4$ temporally adjacent frames before forecasting. 
Since KITTI only provides a current frame and $3$ proceeding frames, we carefully map images back to the raw dataset in order to forecast.
For instance, when $n_f=2$ we infer using frames $[-5, -4, -3, -2]$ then forecast ego-motion and Kalman $n_f$ times.
We then evaluate with respect to frame $0$ which is the standard timestamp KITTI provides images and 3D labels for.
We provide detailed performances on AP$_\three$ in Tab.~\ref{tab:forecast_3D} and AP$_\text{BEV}$ in Tab.~\ref{tab:forecast_BEV}.
We find that the forecasting performance degrades through time but performs reasonably $1-2$ frames ahead, being competitive in magnitude to state-of-the-art methods on the KITTI test dataset as reported in Tab.~$1$ of the main paper. \linebreak
For instance, forecasting $1$ and $2$ frames results in $10.64\%$ and $5.10\%$ AP$_\three$ respectively, which are competitive to methods~\cite{liu2019deep, manhardt2019roi, li2019gs3d, ku2019monocular, simonelli2019disentangling, brazil2019m3d, ma2019accurate} on test. 

\begin{table*}
\begin{center}
\setlength\tabcolsep{4.00pt}
\begin{tabular}{l | c c c  }
\hline
\multirow{2}{*}{ } & \multicolumn{3}{c}{AP$_\three$ \small{(IoU $\geq [0.7/0.5/0.3]$)}}   \\ 
 & Easy & Mod & Hard\\ 
\hline
Forecast $\to 4$ & ~~$1.16$~/~$18.47$~/~$47.26$ & ~~$0.84$~/~$11.21$~/~$29.22$ & ~~$0.62$~/~~$8.97$~/~$23.40$ \\
Forecast $\to3$ & ~~$3.72$~/~$28.97$~/~$58.46$ & ~~$2.32$~/~$18.05$~/~$37.82$ & ~~$1.75$~/~$13.88$~/~$29.80$ \\
Forecast $\to2$ & ~~$7.84$~/~$39.40$~/~$68.87$ & ~~$5.10$~/~$25.48$~/~$48.30$ & ~~$4.14$~/~$20.20$~/~$37.84$ \\
Forecast $\to1$ & $16.09$~/~$49.66$~/~$75.88$ & $10.64$~/~$34.18$~/~$55.26$ & ~~$8.14$~/~$26.62$~/~$44.01$ \\
\hline
No Forecast & ${19.76}$~/~${55.44}$~/~${79.81}$ & ${14.10}$~/~${39.47}$~/~${60.57}$ & ${10.47}$~/~${31.26}$~/~${48.95}$  \\
\hline
\end{tabular}
\end{center}
\caption{
\textbf{Forecasting - 3D Object Detection}.
We evaluate our forecasting performance on AP$_\three$ within the KITTI validation~\cite{chen20153d} set and using IoU$~\geq~0.7/0.5/0.3$.
} \label{tab:forecast_3D}
\vspace{-8mm}
\end{table*}

\begin{table*}
\begin{center}
\setlength\tabcolsep{4.00pt}
\begin{tabular}{l | c c c }
\hline
\multirow{2}{*}{ } & \multicolumn{3}{c}{AP$_\text{BEV}$ \small{(IoU $\geq [0.7/0.5/0.3]$)}}   \\ 
 & Easy & Mod & Hard\\ 
\hline
Forecast $\to 4$ & ~~$5.48$~/~$29.40$~/~$54.52$ & ~~$3.54$~/~$18.13$~/~$36.13$ & ~~$2.90$~/~$14.71$~/~$28.49$ \\
Forecast $\to3$ & $11.03$~/~$39.08$~/~$64.87$ & ~~$6.89$~/~$24.01$~/~$43.52$ & ~~$5.67$~/~$18.85$~/~$34.91$ \\
Forecast $\to2$ & $17.02$~/~$47.07$~/~$72.33$ & $10.76$~/~$31.62$~/~$51.67$ & ~~$8.37$~/~$25.47$~/~$40.79$ \\
Forecast $\to1$ & $23.58$~/~$55.99$~/~$77.48$ & $15.79$~/~$39.33$~/~$58.05$ & $12.54$~/~$31.22$~/~$46.59$ \\
\hline
No Forecast & ${27.83}$~/~${61.79}$~/~${81.20}$ & ${19.72}$~/~${44.68}$~/~${63.44}$ & ${15.10}$~/~${34.56}$~/~${49.84}$ \\
\hline
\end{tabular}
\end{center}
\caption{
\textbf{Forecasting - Bird's Eye View}.
We evaluate our forecasting performance on AP$_\text{BEV}$ within the KITTI validation~\cite{chen20153d} set and using IoU$~\geq~0.7/0.5/0.3$.
} \label{tab:forecast_BEV}
\vspace{-8mm}
\end{table*}

\section{Qualitative Video}
We further provide a qualitative demonstration video at \small\url{http://cvlab.cse.msu.edu/project-kinematic.html}\normalsize. 
The video demonstrates our framework's ability to determine a full scene understanding including 3D object cuboids, per-object velocity and ego-motion. 
We compare to a related monocular work of M3D-RPN~\cite{brazil2019m3d}, plot ground truths, image view, Bird's Eye View, and the track history. 


\clearpage
\bibliographystyle{splncs04}
\bibliography{egbib}

\end{document}